\documentclass[letterpaper, 10 pt, conference]{ieeeconf}  

\IEEEoverridecommandlockouts                              

\title{\LARGE \bf IRisPath: Enhancing Costmap for Off-Road Navigation with Robust IR-RGB Fusion for Improved Day and Night Traversability
}

\author{Saksham Sharma $^{1}$, Akshit Raizada$^{2}$ and Suresh Sundaram$^{3}$
\thanks{$^{1}$Saksham Sharma is with Department of Aerospace Engineering, Indian Institute of Science, India
        {\tt\small saksham.prob@gmail.com}}%
\thanks{$^{2}$Akshit Raizada is a student at the Indian Institute of Technology Indore, India
        {\tt\small me210003009@iiti.ac.in}}%
\thanks{$^{3}$Suresh Sundaram is with the Department of Aerospace Engineering, Indian Institute of Science, India
        {\tt\small vssuresh@iisc.ac.in}}%
}
\let\labelindent\relax

\usepackage{hhline}
\RequirePackage{amssymb}
\RequirePackage{amsmath}
\usepackage{graphicx,subfigure}
\RequirePackage{algpseudocode}
\RequirePackage{algorithm, algorithmicx}
\RequirePackage{mathtools}
\RequirePackage{enumitem}
\RequirePackage{fancyhdr}
\RequirePackage[hidelinks]{hyperref}
\RequirePackage[explicit]{titlesec}
\RequirePackage{lastpage}
\RequirePackage{tikz}
\RequirePackage{pgfplots}
\RequirePackage{xcolor}
\RequirePackage{xparse}
\usepackage[font=small,labelfont=bf,tableposition=top]{caption}

\let\oldtwocolumn\twocolumn
\renewcommand\twocolumn[1][]{%
    \oldtwocolumn[{#1}{
    \begin{center}
           \includegraphics[width=1\linewidth]{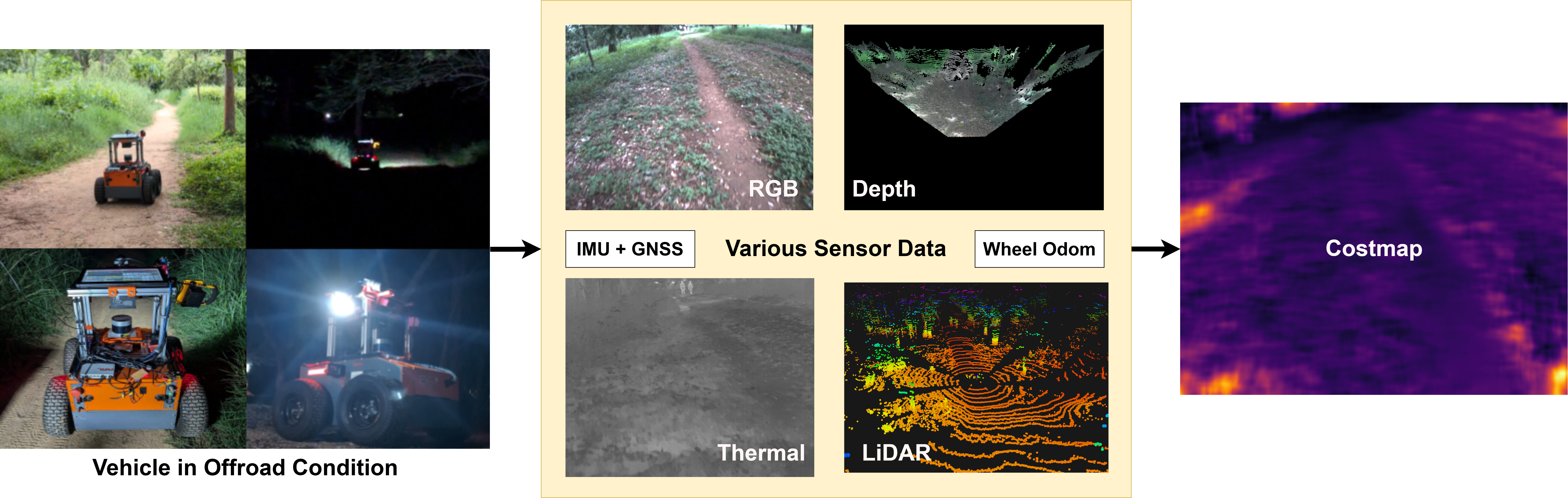}
           \captionof{figure}{A rich multi-modal dataset being collected using Copernicus in an off-road terrain during day and night time for learning traversability costmap of the terrain. Dataset contains LWIR and RGB images along with poincloud from LiDARs and stereo camera.}
           \label{fig:landing_image}
        \end{center}
    }]
}

\begin{document}

\maketitle
\thispagestyle{empty}
\pagestyle{empty}

\begin{abstract}
Autonomous off-road navigation is required for applications in agriculture, construction, search and rescue and defence. Traditional on-road autonomous methods struggle with dynamic terrains, leading to poor vehicle control in off-road conditions. Recent deep-learning models have used perception sensors along with kinesthetic feedback for navigation on such terrains. However, this approach has out-of-domain uncertainty. Factors like change in time of day and weather impacts the performance of the model. We propose a multi modal fusion network "IRisPath" capable of using Thermal and RGB images to provide robustness against dynamic weather and light conditions. To aid further works in this domain, we also open-source a day-night dataset with Thermal and RGB images along with pseudo-labels for traversability. In order to co-register for fusion model we also develop a novel method for targetless extrinsic calibration of Thermal, LiDAR and RGB cameras with translation accuracy of $\pm$1.7cm and rotation accuracy of $\pm$0.827$^{\circ}$. Our dataset and codebase will be available at \href{https://github.com/codeck313/IRisPath}{https://github.com/codeck313/IRisPath}
\end{abstract}

\section{Introduction}
Autonomous navigation in off-road terrain poses a challenge \cite{LeCun2005OffRoadOA, 1641763, 9981942}, and navigating in a robust fashion in different weather and lighting conditions is an even greater problem. The vehicle must navigate these unstructured terrains by analyzing the risk that various paths provide. The affordance of the vehicle is dictated by the vehicle design and so are the risks that various paths may pose. During night time, exteroceptive sensors face significant challenges due to domain shift. RGB cameras experience degraded visibility and diminished semantic information as the visual features undergo substantial changes without sunlight. This environmental transition creates a fundamentally different perceptual landscape compared to day time conditions, complicating reliable feature detection and matching. In addition, common off-road conditions, such as fog and dust, can cause further degradation of the sensors.

Previous approaches \cite{Viswanath2021OFFSEGAS, Guan2021TTMTT, Guan2021GANavET} for off-road traversability extended the semantic methods used for structured environments such as roads. Certain studies \cite{Viswanath2021OFFSEGAS, Guan2021GANavET} have explored discretizing terrain variation into predefined semantic classes, simplifying analysis and representation. Although, it reduces the complexity of the system, discretization also introduces uncertainty, ultimately limiting vehicle performance. Other approaches \cite{Urmson2007} use LiDAR to create occupancy grid maps that capture terrain roughness in a 2.5D elevation map. However, these maps often misclassify small shrubs and plants as obstacles and ignore the robot's state (such as velocity), leading to suboptimal planning performance. Recent methods \cite{pmlr-v164-shaban22a} have also applied learning techniques for direct semantic classification of point clouds. While promising, these approaches require large amounts of hand-annotated data and may overlook subtle terrain nuances due to class discretization.

Self-supervised approaches \cite{9981942, 10160856} have shown great promise in capturing terrain-vehicle interactions while reducing the need for manual labeling. These methods use Inertial Measurement Unit(IMU) feedback and the vehicle's state to assess traversability, adapting to different terrains. However, because these methods learn the correlation between perception, vehicle state, and costmap generation, they struggle with out-of-domain scenarios. Sudden environmental changes, such as dust, fog, or even variations in time of day, can lead to instability, posing significant challenges for real-world deployment.

\begin{table*}[!htb]
\caption{Comparison of various off-road datasets}
\label{tab:datasets_comparision}
\centering
\begin{tabular}{c||c|c|c|p{0.9cm}|p{0.8cm}|p{0.8cm}|c|p{0.8cm}|c|c|c}

Dataset & RGB Image & LWIR Image & Night Time & Stereo Camera & 360 degree Lidar & Solid State Lidar & IMU & Wheel Odom & Calibrated & Action & GPS \\ \hline
 RUGD\cite{Wigness2019ARD}&  Yes&  No& No & No&  Yes&  No&  Yes&  No&  No&  No&  Yes\\ \
 RELLIS-3D\cite{Jiang2020RELLIS3DDD}&  Yes&  No& No & Yes&  Yes&  No&  Yes&  Yes&  Yes&  Yes&  Yes\\ 
 Freiburg Forest\cite{Valada2016DeepMS}&  Yes&  No& No & Yes&  No&  No&  No&  No&  No&  No&  No\\ 
 GOOSE\cite{Mortimer2023TheGD}&  Yes&  No& No & No&  No&  No&  Yes&  No&  Yes&  No&  Yes\\ 
 FinnForest\cite{Ali2020FinnForestDA}&  Yes&  No& No & Yes&  No&  No&  Yes&  No&  Yes&  Yes&  Yes\\ 
 FoMo\cite{Boxan2024FoMoAP}&  Yes&  No& No & Yes&  Yes&  No&  Yes&  Yes&  Yes&  Yes&  Yes\\ 
 TartanDrive 2.0\cite{Sivaprakasam2024TartanDrive2M}&  Yes&  No& No & Yes&  Yes&  Yes&  Yes&  Yes&  No&  Yes&  Yes\\ 
 \textbf{Ours (IRisPath)}&  \textbf{Yes}&  \textbf{Yes}& \textbf{Yes} &  \textbf{Yes}&  \textbf{Yes}&  \textbf{Yes}&  \textbf{Yes}&  \textbf{Yes}&  \textbf{Yes}&  \textbf{Yes}&  \textbf{Yes}\\ 
\end{tabular}
\end{table*}

This paper presents a method to improve self-supervised costmap generation for off-road navigation by addressing domain variations like weather and lighting changes. Our approach leverages sensor fusion between RGB and LWIR (Long-Wave Infrared) imagery to enhance robustness.
RGB cameras provide detailed semantic information but are vulnerable to environmental variations, while LWIR sensors offer superior resilience against these challenges. By combining these complementary modalities, our fusion model \textit{IRisPath} significantly improves the estimation of traversability in diverse conditions. The model architecture efficiently integrates multimodal data to produce reliable costmaps even in challenging scenarios where single-modality approaches typically fail.
To support this research, we introduce a novel dataset featuring co-registered RGB and LWIR images along with synchronized LiDAR and IMU data. Unlike existing Near-Infrared (NIR) light-based datasets \cite{Mortimer2023TheGD, Valada2016DeepMS}, our LWIR data provides greater invariance to lighting conditions. Additionally, we present a targetless calibration technique that enables precise alignment between IR and RGB sensors - addressing a significant barrier to multimodal fusion for off-road navigation. In summary, the contributions of this paper are as follows:
\begin{enumerate}
    \item A robust multi-modal fusion architecture \textit{IRisPath} that leverages both RGB and LWIR imagery to significantly improve traversability estimation in challenging off-road environments with varying weather and lighting conditions.
    \item A comprehensive off-road dataset collected using the Copernicus vehicle, featuring synchronized LWIR and RGB imagery across day and night conditions, complemented by LiDAR and IMU sensor data.
    \item A novel targetless extrinsic calibration methodology that enables accurate registration between LWIR, RGB cameras, and LiDAR sensors.
    \item Experimental results showing the fusion model’s advantages over single modality models.
\end{enumerate}

\section{Related Works}
Terrain traversability assessment combines geometric properties with vehicle characteristics such as stiffness, friction, and granularity. Machine learning approaches for this task gained prominence through initiatives like the DARPA(Defense Advanced Research Projects Agency) LAGR(Learning Applied to Ground Vehicles) \cite{lagr} and DARPA Grand Challenge \cite{Behringer2004TheDG} programs. LeCun \textit{et al.} \cite{LeCun2005OffRoadOA} pioneered the use of Convolutional Neural Networks (CNNs) for off-road navigation. Early work \cite{Wigness2019ARD, Jiang2020RELLIS3DDD} primarily focused on semantic segmentation of predefined terrain classes, while Shaban \textit{et al.} \cite{pmlr-v164-shaban22a} extended this by directly generating cost maps from LiDAR-based geometric data. However, these discrete segmentation methods provide limited adaptability in complex off-road environments.

More recently, self-supervised approaches \cite{9981942,10160856}, inspired by earlier works \cite{Wellhausen2019WhereSI, 8460731}, have emerged to generate traversability costmaps using IMU data. While effective, these methods struggle with robustness in varying weather and lighting conditions. To evaluate and improve such navigation models, datasets covering both day and night scenarios are essential. RGB imagery alone is highly sensitive to environmental changes \cite{9879642}, limiting its reliability. In contrast, integrating LWIR with RGB can significantly enhance perception, ensuring robustness to varying lighting conditions.

Despite the growing demand for off-road datasets, existing datasets have limitations. Robot Unstructured Ground Driving (RUGD) \cite{Wigness2019ARD} and RELLIS-3D \cite{Jiang2020RELLIS3DDD} offer multi-modal data but require extensive manual annotation, which can introduce inconsistencies in dynamic environments. Forêt Montmorency (FoMo) \cite{Boxan2024FoMoAP} provides multi-season data from boreal forests. TartanDrive 2.0 \cite{Sivaprakasam2024TartanDrive2M}, GOOSE \cite{Mortimer2023TheGD}, Freiburg-Forest \cite{Valada2016DeepMS} captures diverse off-road dynamics through an extensive sensor suite. However, none of these datasets include pure night time data or LWIR imagery, limiting their applicability for robust, all-condition navigation. Table \ref{tab:datasets_comparision} summarizes these datasets.

For enabling fusion algorithms in navigation, accurate calibration between LWIR and RGB cameras is essential. This ensures precise alignment of both modalities, allowing for robust perception across varying conditions. Existing calibration methods fall into two categories: target-based, which require predefined calibration patterns \cite{Dhall2017LiDARCameraCU, Fang2021SingleShotIE}, and targetless methods, which estimate camera motion \cite{camera_motion} or leverage deep learning to extract semantic information for feature matching \cite{Luo2023CalibAnythingZL, Ye2024MFCalibSA}. While targetless approaches have been explored for RGB cameras, their application to thermal cameras remains largely unexplored. Most thermal calibration methods rely on specially designed targets that are visible in both thermal and RGB spaces \cite{Li2018SpatialCF, s23073479}. However, recent interest has shifted toward feature-based techniques for real-time calibration. Fu \textit{et al.} \cite{Fu2021TargetlessEC} demonstrated an approach using edge features for stereo camera and LiDAR calibration, though it relies solely on geometric features and does not utilize LiDAR reflectivity to enhance accuracy. Given the importance of LWIR in night time navigation, the ability to perform on-the-fly calibration is increasingly critical for ensuring seamless sensor fusion.

\section{IRIsPath: Robust Costmap using Fusion}

Our IRIsPath network predicts vehicle dynamics dependent traversability cost maps using RGB and LWIR camera. RGB images provide rich semantic information in well-lit environments but become ineffective in low-light or challenging weather conditions. LWIR images offer resilience in these adverse conditions despite their lower resolution. Our proposed model fuses both modalities to leverage their complementary strengths—RGB's detailed visual information and LWIR's consistent performance in low-visibility scenarios—creating a robust system capable of reliable costmap generation across diverse environmental settings. 

In the following sections we define our vehicle setup, calibration process, and our fusion based costmap model.

\subsection{Hardware Setup}
\begin{figure}[!htb]
    \centering
    \includegraphics[width=0.5\textwidth]{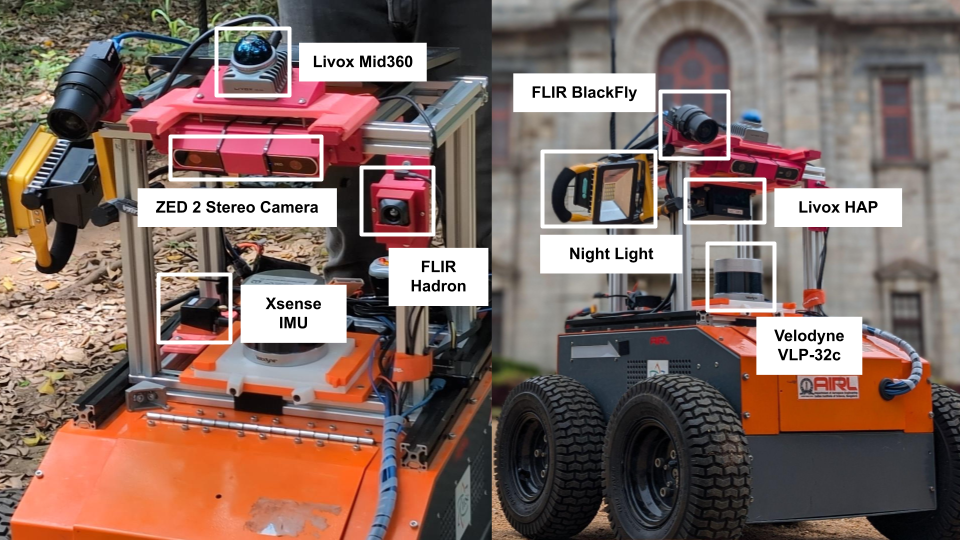}
    \caption{Various sensors mounted on Copernicus for the off-road dataset collection.}
    \label{fig:sensor_setup}
\end{figure}
We employ a customized version of the all-terrain ground robot Copernicus, developed by BotSync, to collect the data as seen in Fig. \ref{fig:sensor_setup}. The robot is equipped with the following sensors:

\begin{enumerate}
    \item \textit{FLIR's Hadron Camera}: A LWIR camera, well suited for low-visibility conditions, such as fog, dust, or night time. 
    \item \textit{FLIR's BlackFly S Camera}: It is the main RGB sensor (2k resolution) tilt mounted on the vehicle.
    \item \textit{ZED-2 Stereo Camera}: A depth camera that captures surface roughness.
    \item \textit{Livox's Mid360 LiDAR}: Non-repeating 360-degree LiDAR tilt mounted to capture the terrain ahead of the vehicle.
    \item \textit{Velodyne 32C LiDAR}: Repeating 360-degree LiDAR is mounted horizontally on the vehicle in order to sense any nearby obstacle.
    \item \textit{Livox's HAP LiDAR}: An automotive-grade solid-state LiDAR complements the Velodyne by offering precise data directly ahead of the vehicle.
    \item \textit{Xsense MTI-680G IMU}: High accuracy IMU along with GNSS for global positioning.
    \item \textit{Wheel RPM Sensors}: Enable wheel odometry for vehicle motion estimation.
\end{enumerate}

\subsection{Calibration}

\begin{figure}
    \centering
    \includegraphics[width=0.5\textwidth]{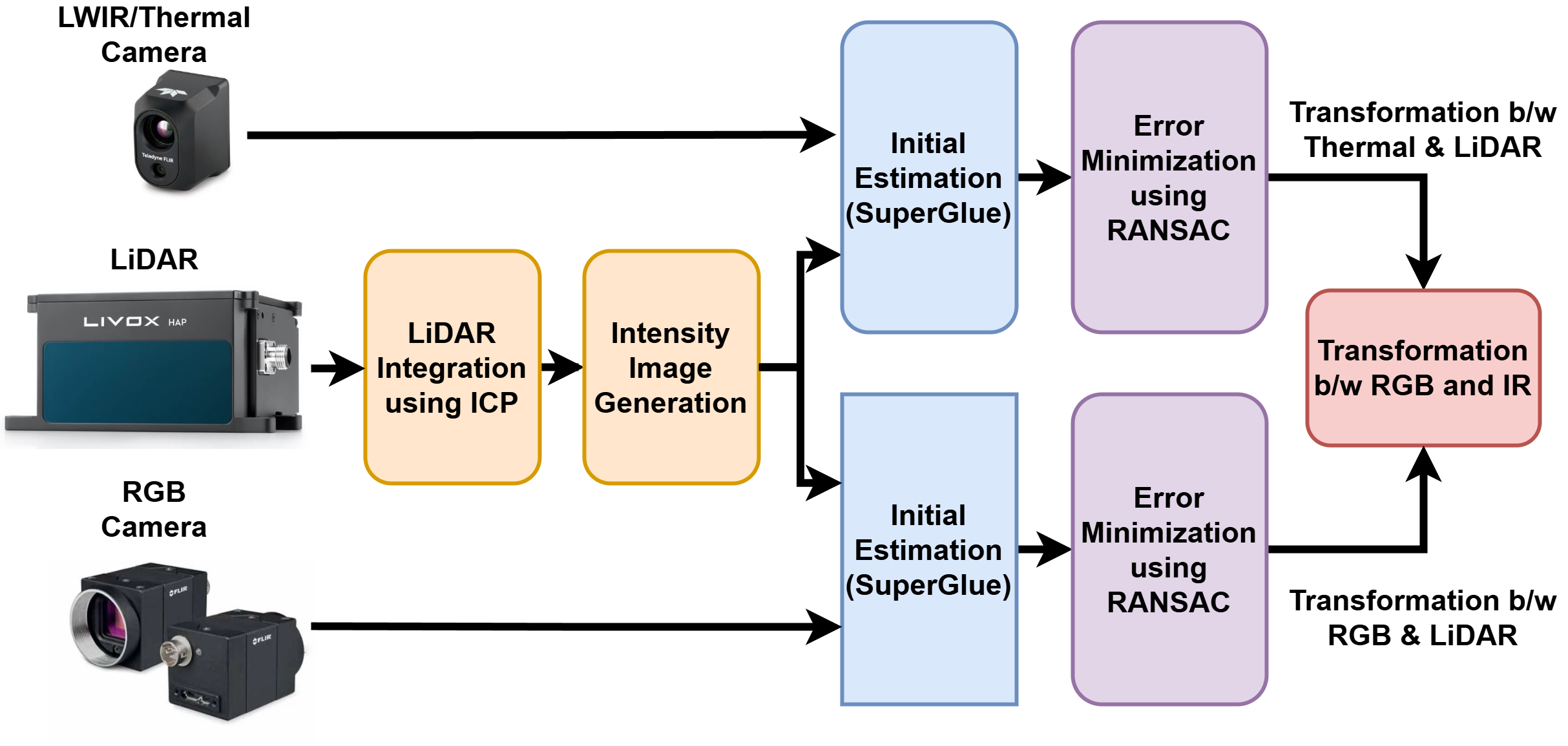}
    \caption{IR-RGB Calibration Process to obtain extrinsic transformation parameters}
    \label{fig:calib_process}
\end{figure}

To integrate both IR and RGB modalities in our model, extrinsic calibration between the two cameras is essential. Co-registering IR and RGB images is challenging, as it typically requires specialized checkerboards. In off-road environments, sensor displacement is common, necessitating frequent recalibration. Target-based methods are impractical for regular use, so we developed a targetless extrinsic calibration approach, using LiDAR as an intermediary. LiDAR retains structural features visible in the RGB modality while being semantically closer to the IR modality, making it a suitable bridge between the two. Given the lack of robust feature-matching models for RGB-IR calibration, this approach offers a practical compromise, especially since the system will operate with LiDAR onboard.

The calibration setup (as seen in Fig. \ref{fig:calib_process}) involves two cameras and LiDAR that have a shared field of view (FOV) as seen in Fig. \ref{fig:sensor_setup}. Our goal is to estimate the transformation from RGB to IR frame, $^{rgb}T_{ir}$. We first generate a dense point cloud using a non-repeating LiDAR or a repeating LiDAR with Iterative Closest Point (ICP)-based mapping.

This point cloud is then converted into a LiDAR intensity image. Using SuperGlue or ORB, we extract common features between the LiDAR intensity image and both IR and RGB images to compute transformations $^{ir}T_{li}$ and $^{rgb}T_{li}$.

A point $x_{rgb}$ in RGB frame is projected to world space as $w_{rgb} = \pi^{-1}(x_{rgb})$, where $\pi$ is the projection function, then transformed to the LiDAR frame as $w_{li} = ^{rgb}T_{li} w_{rgb}$. Similarly, the IR image point can be transformed to LiDAR frame using $w_{li} = ^{ir}T_{li} \pi^{-1}(x_{ir})$.

\begin{figure*}[!htb]
    \centering
    \includegraphics[width=\linewidth]{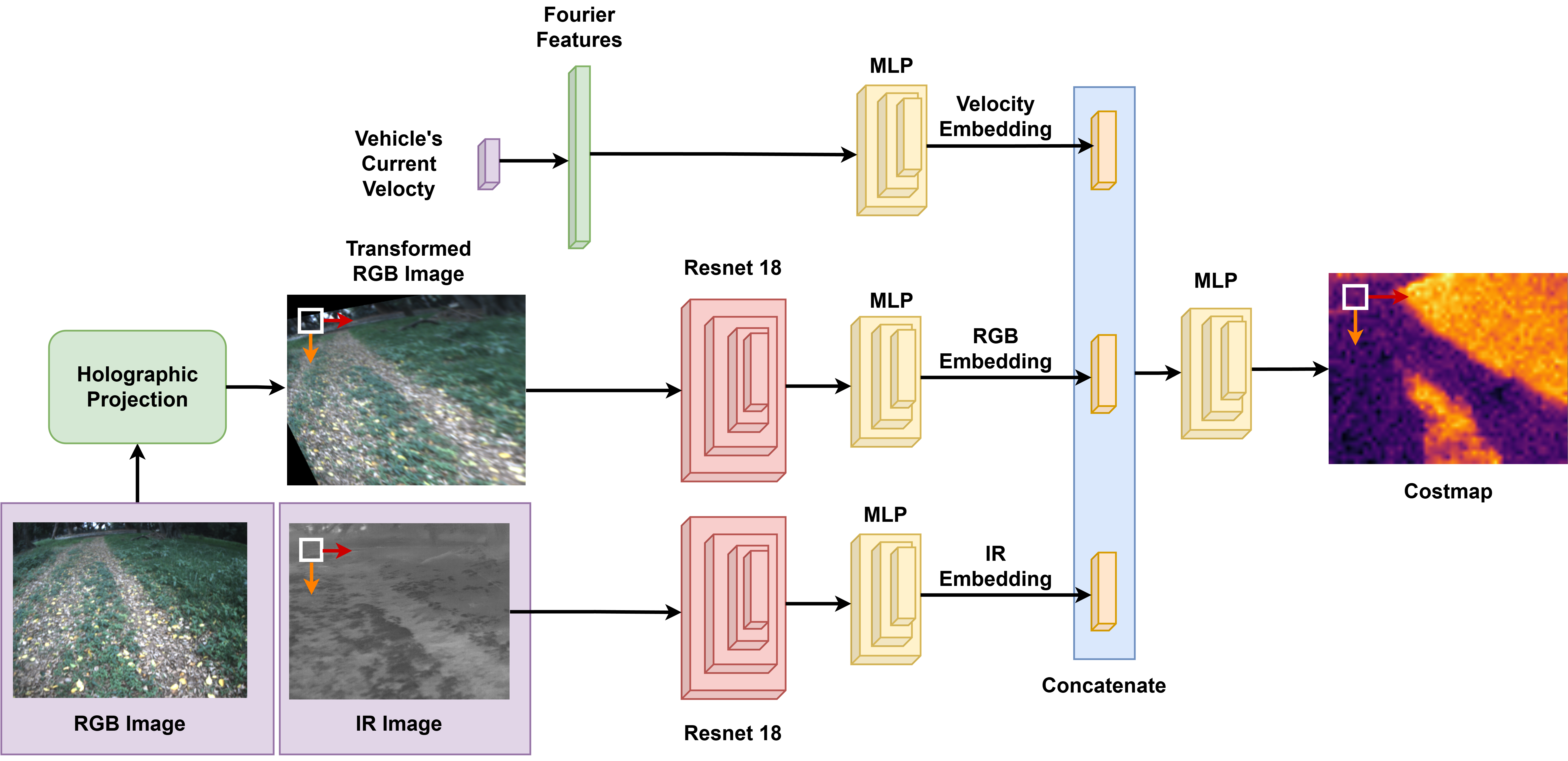}
    \caption{Overview of the proposed early fusion approach. The inputs to the model are in the purple boxes. To get both the input image in the same frame we apply homographic projection onto the RGB image so that we get matching image pairs of RGB and IR. Then we first split both input images into multiple smaller images on which model predicts a traversability score which are then combined and shown as a costmap.}
    \label{fig:cnn_archi}
\end{figure*}

To find the image to LiDAR transformation, we use Random Sample Consensus (RANSAC) and re-projection error minimization. Consider RGB image's keypoints to be $[x_{rgb, 1},x_{rgb, 2},...,x_{rgb, N}]$ and LiDAR intensity image's keypoints to be $[w_1,w_2,...,w_N]$. We can find the $^{rgb}T_{li}$ by minimizing,
\begin{equation} \label{eq:1}
\arg \min_{^{rgb}T_{li}} \sum_{i=1}^{N} || w_i - ^{ir}T_{li} \: (\pi^{-1}(x_{rgb,i})) ||^2
\end{equation}

Once we have both $^{rgb}T_{li}$ and $^{ir}T_{li}$ we can calculate,
\begin{equation} \label{eq:2}
^{rgb}T_{ir} = \, ^{rgb}T_{li} \: (^{ir}T_{li})^{-1} 
\end{equation}

Using these transformation matrices, the RGB image is aligned with the IR camera's perspective. The RGB image is transformed instead of the IR image due to its higher resolution, ensuring minimal data degradation.

\subsection{Robust Traversability Costmap using Fused Modalities}
In this section we describe our fusion model pipeline. Overview of this pipeline can be seen in Fig. \ref{fig:cnn_archi}
\subsubsection{Preprocessing Inputs}
We first apply a transformation matrix to co-register the RGB and IR images, aligning them spatially to mitigate perspective distortions that could degrade model performance.

Inspired by TerraPN \cite{9981942}, we sample sub-images from the co-registered RGB-IR pair at test time, with the model computing a traversability cost for each. This converts the problem into a regression task, where each sub-image is assigned a cost value, restructured into a cost grid matching the original image dimensions. Unlike TerraPN, our method allows flexible output resolution by adjusting stride and input size, optimizing for available computational resources.

During training, instead of resizing the entire image like TerraPN, we extract only patches beneath the robot, ensuring IMU data corresponds directly to the terrain underfoot rather than ahead. For example, if a rock causes an IMU spike, our approach associates that reading with the exact image patch beneath the robot at that moment. This direct mapping improves the model’s ability to learn terrain-traversability relationships.

For an input image of width $w$ and height $h$, with sampled patches of size $i \times i$ and stride $s$, the number of generated image pairs is: 
\begin{equation} \label{eq:ff}
numImagePairs = \lfloor \frac{w-i}{s} + 1 \rfloor \times \lfloor \frac{h-i}{s} + 1 \rfloor
\end{equation}

\subsubsection{Model Architecture}
The model takes three inputs: ego vehicle velocity, a transformed RGB image, and an IR image. The RGB and IR images pass through separate ResNet-18 backbones, extracting modality-specific features, which are then processed by independent multi-layer perceptrons (MLPs) to generate embeddings. This separation preserves complementary information from both modalities.

Velocity is encoded using Fourier features \cite{Tancik2020FourierFL}, which increases the input velocity's dimensionality and enhances high-frequency patterns in low-dimensional data. This approach helps improve costmap predictions. Given the vehicle's velocity norm $v$, we apply Fourier encoding:
\begin{equation} \label{eq:ff}
\gamma(v) = [cos(2\pi Bv), sin(2\pi Bv)]^T
\end{equation}
where $B \in \mathbb{R}^{m\times d}$ is sampled from $\mathcal{N}(0, \sigma^2)$, with $\sigma$ determined via hyperparameter tuning.

The RGB, IR, and velocity embeddings are concatenated and processed through an MLP, allowing the model to learn non-trivial correlations between modalities. This fusion improves robustness across diverse terrains and environmental conditions by leveraging their combined strengths.

\subsubsection{Cost Function}

Our model operates in a self-supervised manner, using a loss function that incorporates the robot's z-axis acceleration—specifically, the frequency of vertical perturbations—as a cost metric for each image patch. We compute the area under the power spectral density (PSD) of acceleration and normalize it with the vehicle's velocity to account for momentum:

\begin{equation} \label{eq:3}
y = \frac{PSD(acc_z)}{\sqrt{V_x^2 + V_y^2}+10}
\end{equation}
where $y$ is the traversability cost used as a label, $PSD(acc_z)$ is the power spectral density of z-axis acceleration, $V_x$ and $V_y$ are the vehicle’s velocities. The normalization prevents disproportionately low cost scores at slow speeds, where cautious driving (e.g., on grass) might otherwise misrepresent terrain difficulty. The added $10$ in the denominator prevents division by near-zero values when the vehicle is stationary. This self-supervised approach improves adaptability and enables precise costmaps without manual labeling.




\section{Experiments and Results}
\subsection{IRIsPath Dataset}
The dataset was collected in an off-road area at the Indian Institute of Science (IISc) in Bangalore (13°01'05.4"N 77°34'02.0"E). This location features diverse terrain including dirt paths, rocky sections, vegetation patches, inclines, and obstacles like boulders and tree roots. We selected pre-defined paths to ensure uniform sampling across this varied environment, providing realistic conditions for ground vehicle testing.  The dataset has a night-to-morning image ratio of 1.79, with further statistics in Table \ref{tab:dataset_specs}. 


\begin{table}[]
\centering
\caption{IRIsPath Dataset Statistics}
\begin{tabular}{|l|l|l|l|}
\hline
Time of Day & RGB Images & IR Images & Total Images \\ \hline
Morning     & 10,555     & 37,620    & 48,175       \\ \hline
Evening     & 18,922     & 73,331    & 92,253       \\ \hline
\end{tabular}
\label{tab:dataset_specs}
\end{table}

\begin{figure}[!htb]
    \centering
    \includegraphics[width=0.5\textwidth]{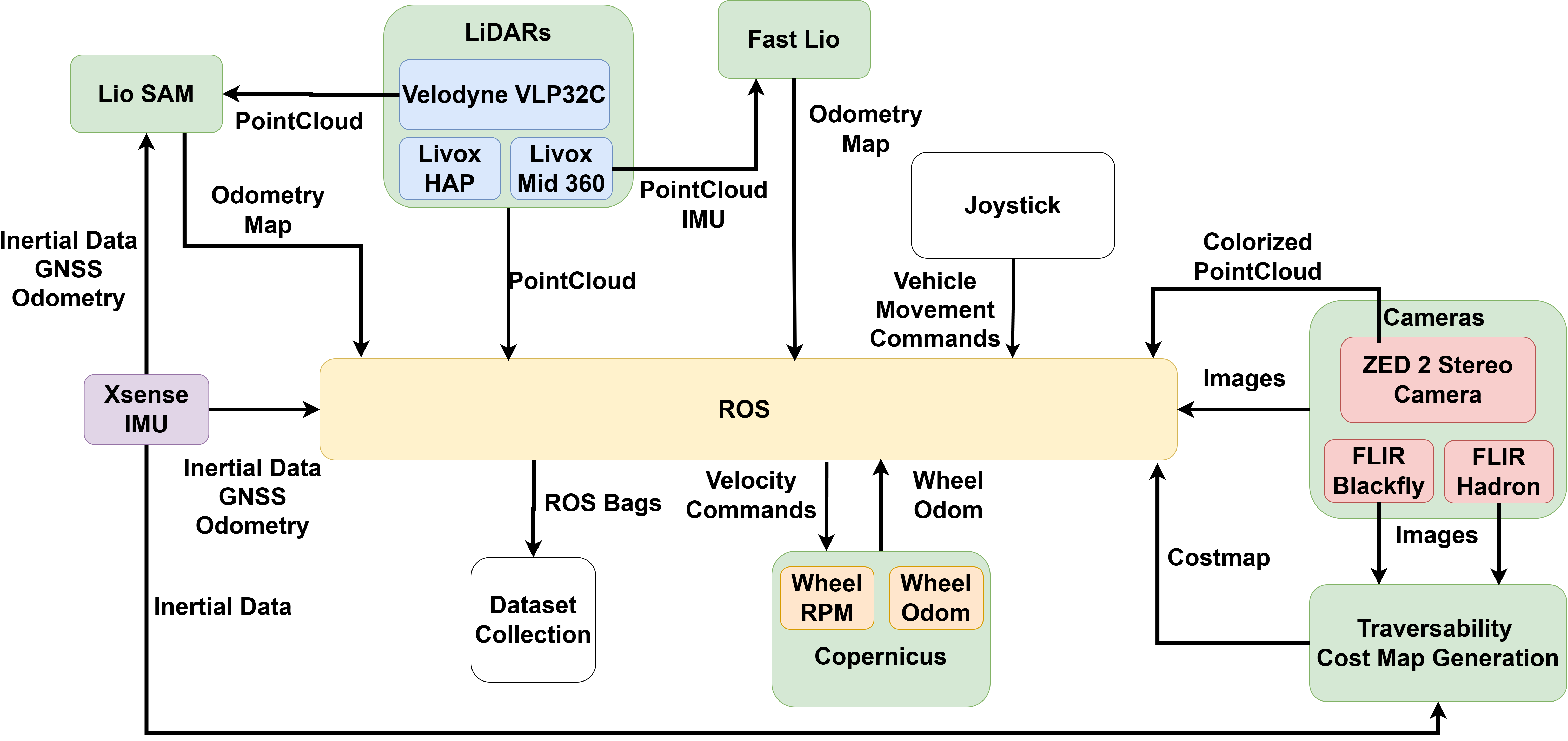}
    \caption{System configuration for dataset collection on the Copernicus vehicle. We initially record all the sensors and actuators as a rosbag and further process them to other formats.}
    \label{fig:setup_dataset}
\end{figure}

In order to aid the user of the dataset for the navigation tasks we post processed data to include odometry from FastLIO2 \cite{fastlio} with Scan-Context++ to provide loop closure for added accuracy of the map. It also provides us with a map of the surroundings using the Mid 360 LiDAR and takes advantage of the non-repeating mode available in Mid360 to improve the point density of the map. We also provide odometry from LIO-SAM \cite{liosam} running with VLP-32 and Global Navigation Satellite System (GNSS) to provide a tightly coupled odometry, enabling highly accurate vehicle trajectory. Fig. \ref{fig:setup_dataset} shows the overall setup and integration of other modules in the dataset.

The dataset is originally recorded as a bagfile. But we also post-process the dataset and convert it to KITTI \cite{Geiger2013IJRR} and NuScenes \cite{nuscenes} format to enable a broader set of applications for our dataset. We have also labelled each run with the following metadata to allow further insight.
\begin{itemize}
    \item Time of data collection (in Indian Standard Time)
    \item Calibration data for all the sensors
    \item Driver and Route ID
    \item Temperature, weather and lighting
\end{itemize}

\subsection{Extrinsic Calibration}
To evaluate our calibration technique, we collected data from various spots inside the lab using a vehicle equipped with RGB camera, IR camera, and LiDAR. Lab objects served as targetless features for calibration. The LiDAR intensity image is used with SuperGlue \cite{9157489} to find initial correspondence estimates. 

\begin{figure}
\centering     
\includegraphics[width=0.4\textwidth]{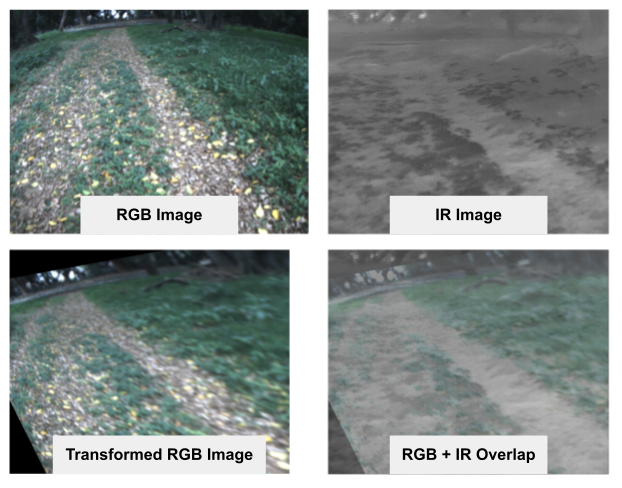}
\caption{The RGB image is being transformed into the IR image frame using the calculated extrinsic parameters.}
\label{fig:calibration}
\end{figure}

We further minimized re-projection error between both cameras and LiDAR  via RANSAC. We calculate $^{rgb}T_{ir}$ from the extrinsic parameter $^{rgb}T_{li}$ and $^{ir}T_{li}$. Table \ref{tab:calib_results} compares our calculated transforms against ground truth measurements obtained using the lab's physical measurement tools. Fig. \ref{fig:calibration} showing RGB camera's image translated to IR camera's pose.

\begin{table}[!thb]
\centering
\caption{Extrinsic RGB to IR Calibration Result}
\label{tab:calib_results}
\begin{tabular}{|l|l|l|l|l|}
\hline
      & Measured & Initial Estimate & Final Estimate & Error    \\ \hline
x     & 7 cm     & 4.0548 cm        & 5cm            & -2cm     \\ \hline
y     & 36 cm    & 29.945 cm        & 34.3 cm        & -1.7cm   \\ \hline
z     & -13 cm   & -27.797 cm       & -11.6cm        & -1.4cm   \\ \hline
roll  & 0$^{\circ}$    & 4.02956$^{\circ}$      & -0.2$^{\circ}$       & 0.2$^{\circ}$   \\ \hline
pitch & -16$^{\circ}$  & -19.12204$^{\circ}$    & -18.11$^{\circ}$     & 2.11$^{\circ}$  \\ \hline
yaw   & 0$^{\circ}$    & -0.61564$^{\circ}$     & 0.171$^{\circ}$      & 0.171$^{\circ}$ \\ \hline
\end{tabular}
\end{table}

\subsection{Qualitative Analysis of the Single Modality Model}

\begin{figure}
    \centering
    \subfigure[At low velocity]{\label{fig:check:lowvel}\includegraphics[width=0.225\textwidth]{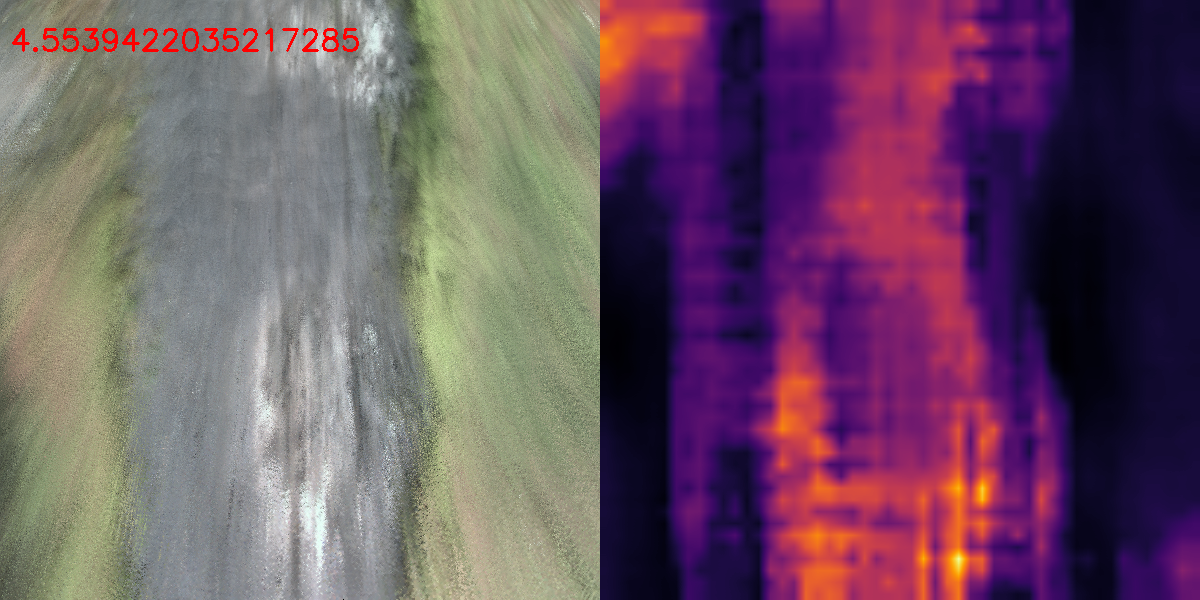}
    } 
    \subfigure[At high velocity]{\label{fig:check:highvel}\includegraphics[width=0.225\textwidth]{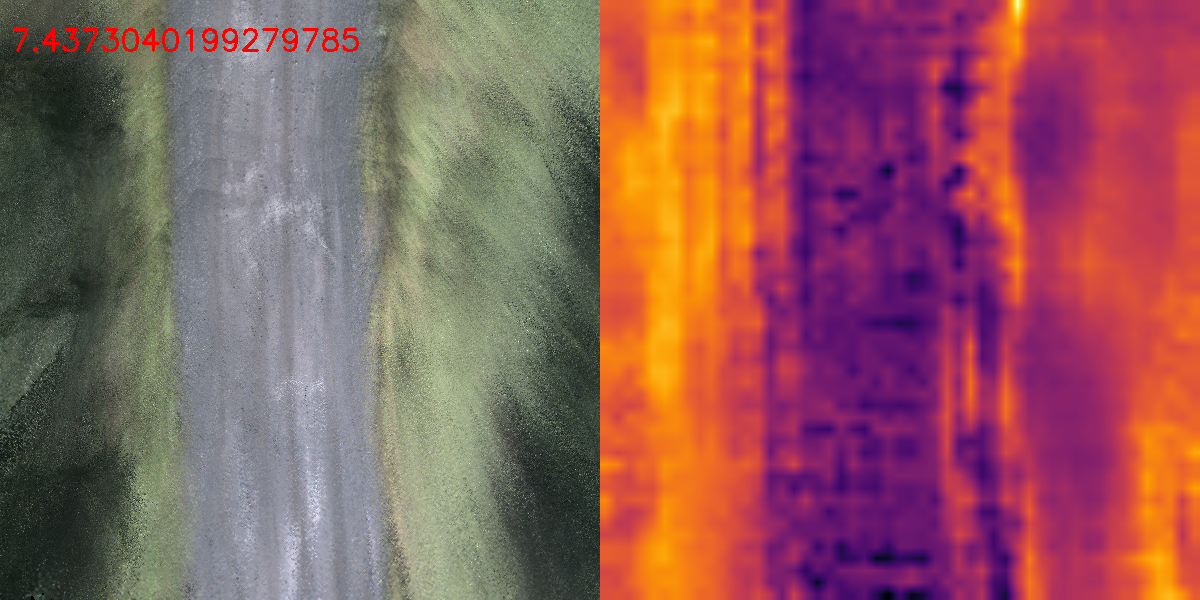}
    } 
    \subfigure[Using less Fourier Features]{\label{fig:check:lessff}\includegraphics[width=0.225\textwidth]{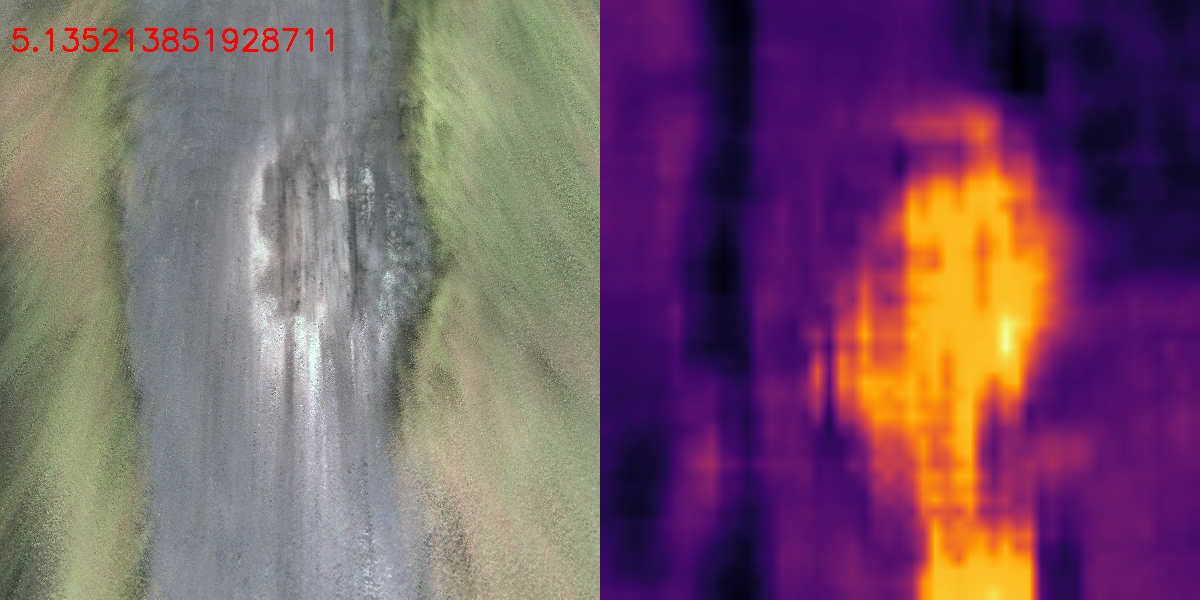}}
    \subfigure[Without Fourier Features]{\label{fig:check:noff}\includegraphics[width=0.225\textwidth]{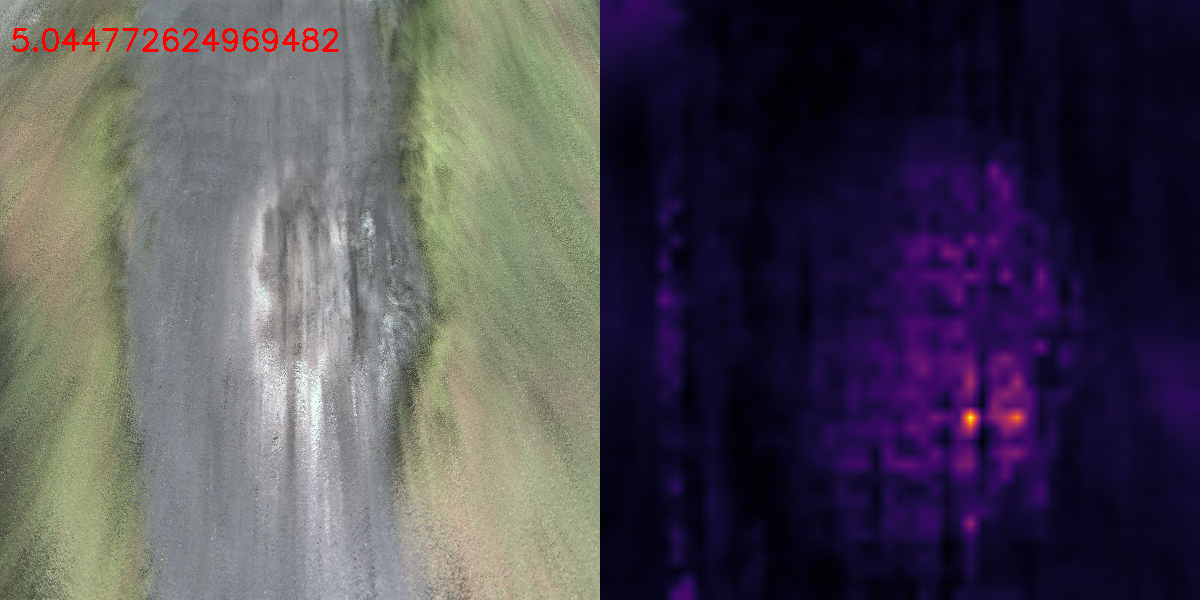}}
    \caption{Qualitative analysis of the model's output with respect to the input velocity and fourier features used in the model.}
    \label{fig:check}
\end{figure}

To evaluate the impact of our velocity-dependent cost function and Fourier features, we conducted an ablation study. The costmaps are color-coded, with higher traversability costs shown in yellow and lower costs in blue. The results are as follows:
\begin{enumerate}
    \item \textbf{Effect of Speed on Costmap} -  Fig. \ref{fig:check:lowvel} and Fig. \ref{fig:check:highvel} shows how speed influences the costmap. At higher speeds (Fig. \ref{fig:check:highvel}), low-impulse costs (e.g., small pebbles) disappear, while regions like grass become high-cost due to terrain uncertainty. Speed also makes the costmap momentum-aware, adjusting lethality based on ego’s velocity—objects that pose minimal risk at low speeds may become hazardous at higher speeds.
    \item \textbf{Fourier Features} - Fig. \ref{fig:check:lowvel} illustrates that Fourier features enable the network to capture high-frequency data from low-dimensional sensors like the IMU. As Fourier features are reduced (Fig. \ref{fig:check:lessff}), regional distinctions in the costmap become less prominent. Without them (Fig. \ref{fig:check:noff}), the costmap fails to differentiate between regions. 
\end{enumerate}


\subsection{Qualitative Analysis of the Fusion Model} 
\begin{figure}
    \centering
    \subfigure[RGB Inference: Day Time]{\label{fig:RGB_IR Inferences:day_rgb}\includegraphics[width=0.245\textwidth]{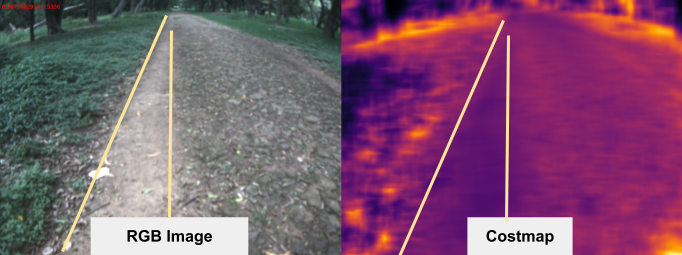}} 
    \subfigure[IR Inference: Day Time]{\label{fig:RGB_IR Inferences:day_ir}\includegraphics[width=0.228\textwidth]{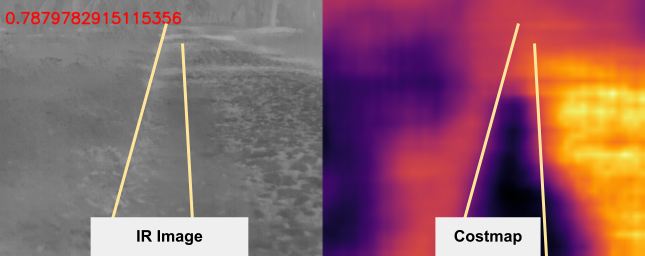}} 
    \subfigure[RGB Inference: Night Time]{\label{fig:RGB_IR Inferences:night_rgb}\includegraphics[width=0.245\textwidth]{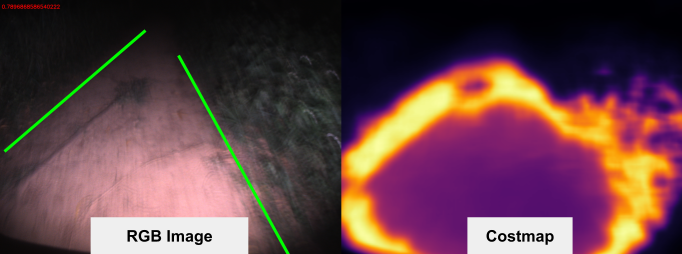}}
    \subfigure[IR Inference: Night Time]{\label{fig:RGB_IR Inferences:night_ir}\includegraphics[width=0.228\textwidth]{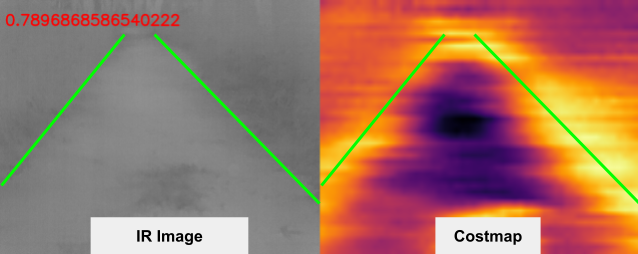}}
    \caption{Costmaps generated by using only single modality in day and night time}
    \label{fig:RGB_IR Inferences}
\end{figure}

\begin{figure}[!htb]
    \centering
    \subfigure[Day time inference using fusion model]{\label{fig:Fusion Inferences:day} \includegraphics[width=0.5\textwidth]{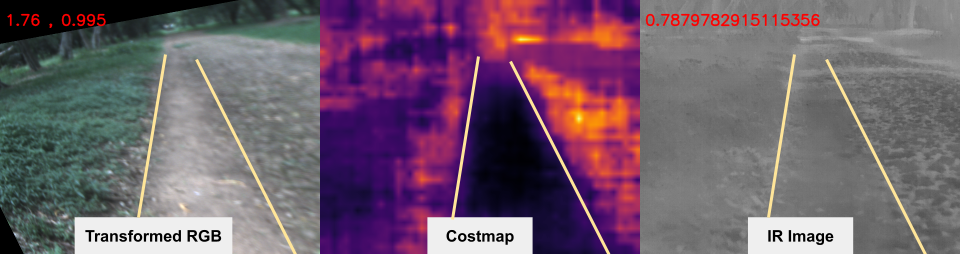}} 
    \subfigure[Night time inference using fusion mode]{\label{fig:Fusion Inferences:night}\includegraphics[width=0.5\textwidth]{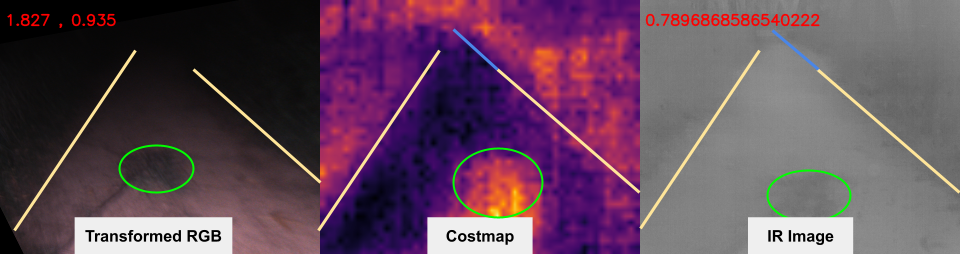}} 
    \caption{Costmaps using fused modality in day and night time.}
    \label{fig:Fusion Inferences}
\end{figure}

To evaluate the fusion method, we infer costmaps using single modalities (RGB or IR) in both day and night conditions (Fig. \ref{fig:RGB_IR Inferences}) and compare them to fusion-based costmaps (Fig. \ref{fig:Fusion Inferences}).

In daylight, RGB-only costmaps (Fig. \ref{fig:RGB_IR Inferences:day_rgb}) are more detailed than IR-only costmaps (Fig. \ref{fig:RGB_IR Inferences:day_ir}). However, at night, RGB-only costmaps (Fig. \ref{fig:RGB_IR Inferences:night_rgb}) fail to capture terrain correctly, whereas IR-only costmaps retain terrain information but appear blurred due to lower resolution and reduced semantic detail.

The fusion model significantly enhances costmap quality in both scenarios. During the day (Fig. \ref{fig:Fusion Inferences:day}), it assigns lower costs to traversable paths than RGB-only costmaps and defines boundaries more precisely than IR-only costmaps.  At night (Fig. \ref{fig:Fusion Inferences:night}), it effectively extracts complementary features from both modalities. While the RGB camera has limited visibility at night, the IR camera captures features further down the road, such as the blue boundary missed by the RGB-only costmap. Meanwhile, features like grass patches and tree roots are more clearly defined in RGB modality. The fused costmap successfully incorporates both, demonstrating the model’s ability to leverage multiple sensors for a more accurate environmental representation.


\section{Conclusion and Future Works}

In this work, we propose the IRIsPath fusion model to enhance day and night traversability by integrating multi-modal visual data and vehicle state feedback into costmap generation. Our fusion model and cost function significantly enhanced the costmap quality compared to single-modality approaches. To enable this fusion, we introduce a novel targetless method for RGB-LWIR extrinsic calibration. Furthermore, we provide a multi-modal day and night dataset to enable future work in offroad navigation. For future work we would investigate using our model as an encoder for an end-to-end navigation system, allowing the system to learn dynamics of the vehicle.




\bibliographystyle{unsrt}
\bibliography{IEEEabrv}

\end{document}